\documentclass[conference,compsoc]{IEEEtran}

\IEEEoverridecommandlockouts
\usepackage{cite}
\usepackage{amsmath,amssymb,amsfonts}
\usepackage{algorithmic}
\usepackage{graphicx}
\usepackage{textcomp}
\usepackage{lipsum}
\usepackage{hyperref}
\usepackage{subcaption}
\usepackage{xcolor}
\usepackage{booktabs}
\usepackage{adjustbox}
\usepackage{booktabs}
\usepackage{multirow}
\usepackage{float}
\usepackage{xspace} 
\begin{document}

\title{ Soft Sensing Model Visualization: \\ Fine-tuning Neural Network from What Model Learned}

\makeatletter
\newcommand{\linebreakand}{%
  \end{@IEEEauthorhalign}
  \hfill\mbox{}\par
  \mbox{}\hfill\begin{@IEEEauthorhalign}
}
\makeatother
\author{
  \IEEEauthorblockN{Xiaoye Qian}
  \IEEEauthorblockA{\textit{Seagate Technology, MN, US} \\
    \textit{Case Western Reserve University, OH, US}\\
    xiaoye.qian@seagate.com}
    \and 
  \IEEEauthorblockN{Chao Zhang}
  \IEEEauthorblockA{\textit{Seagate Technology, MN, US} \\
    \textit{University of Chicago, IL, US}\\
    chao.1.zhang@seagate.com}
  \and
  \IEEEauthorblockN{Jaswanth Yella}
  \IEEEauthorblockA{\textit{Seagate Technology, MN, US} \\
    \textit{University of Cincinnati, OH, US}\\
     jaswanth.k.yella@seagate.com}
  \linebreakand
  \IEEEauthorblockN{Yu Huang}
  \IEEEauthorblockA{\textit{Seagate Technology, MN, US} \\
    \textit{Florida Atlantic University, FL, US}\\
    yu.1.huang@seagate.com}
  \and
  \IEEEauthorblockN{Ming-Chun Huang}
  \IEEEauthorblockA{
    \textit{Duke Kunshan University}\\
    mh596@duke.edu}
  \and
  \IEEEauthorblockN{Sthitie Bom}
  \IEEEauthorblockA{\textit{Seagate Technology} \\
    \textit{MN, US}\\
    sthitie.e.bom@seagate.com}
}

\IEEEoverridecommandlockouts
\IEEEpubid{\makebox[\columnwidth]{This paper has been accepted by 2021 IEEE International Conference on Big Data \hfill} \hspace{\columnsep}\makebox[\columnwidth]{ }}

\maketitle

\begin{abstract}

The growing availability of the data collected from smart manufacturing is changing the paradigms of production monitoring and control. The increasing complexity and content of the wafer manufacturing process in addition to the time-varying unexpected disturbances and uncertainties, make it infeasible to do the control process with model-based approaches. As a result, data-driven soft-sensing modeling has become more prevalent in wafer process diagnostics. Recently, deep learning has been utilized in soft sensing system with promising performance on highly nonlinear and dynamic time-series data. Despite its successes in soft-sensing systems, however, the underlying logic of the deep learning framework is hard to understand. In this paper, we propose a deep learning-based model for defective wafer detection using a highly imbalanced dataset. To understand how the proposed model works, the deep visualization approach is applied. Additionally, the model is then fine-tuned guided by the deep visualization. Extensive experiments are performed to validate the effectiveness of the proposed system. The results provide an interpretation of how the model works and an instructive fine-tuning method based on the interpretation.

\end{abstract}


\IEEEpeerreviewmaketitle

\section{Introduction}

The manufacturing process is becoming increasingly more complex and longer, resulting in a significant increase in cost and difficulty of measuring the key quality variables. \cite{chen2017fault}. In order to obtain effective quality indicators, the industry has developed soft-sensing technology to estimate and predict the quality variables \cite{jiang2020review}. The soft-sensing technology has been widely applied to various industries for real-time quality monitoring and early reports before time-consuming laboratory analysis, including chemical production \cite{poerio2018frequency,yin2014robust}, drug industry \cite{ji2012recursive}, petroleum refining \cite{bidar2017data}, mechanical industry \cite{ding2012novel}, fault detection and diagnosis \cite{downs1993plant, mattera2018method} and semiconductor manufacturing industry \cite{fan2020defective}. 

\begin{figure}[t]
	\centering
	\includegraphics[width=0.48\textwidth]{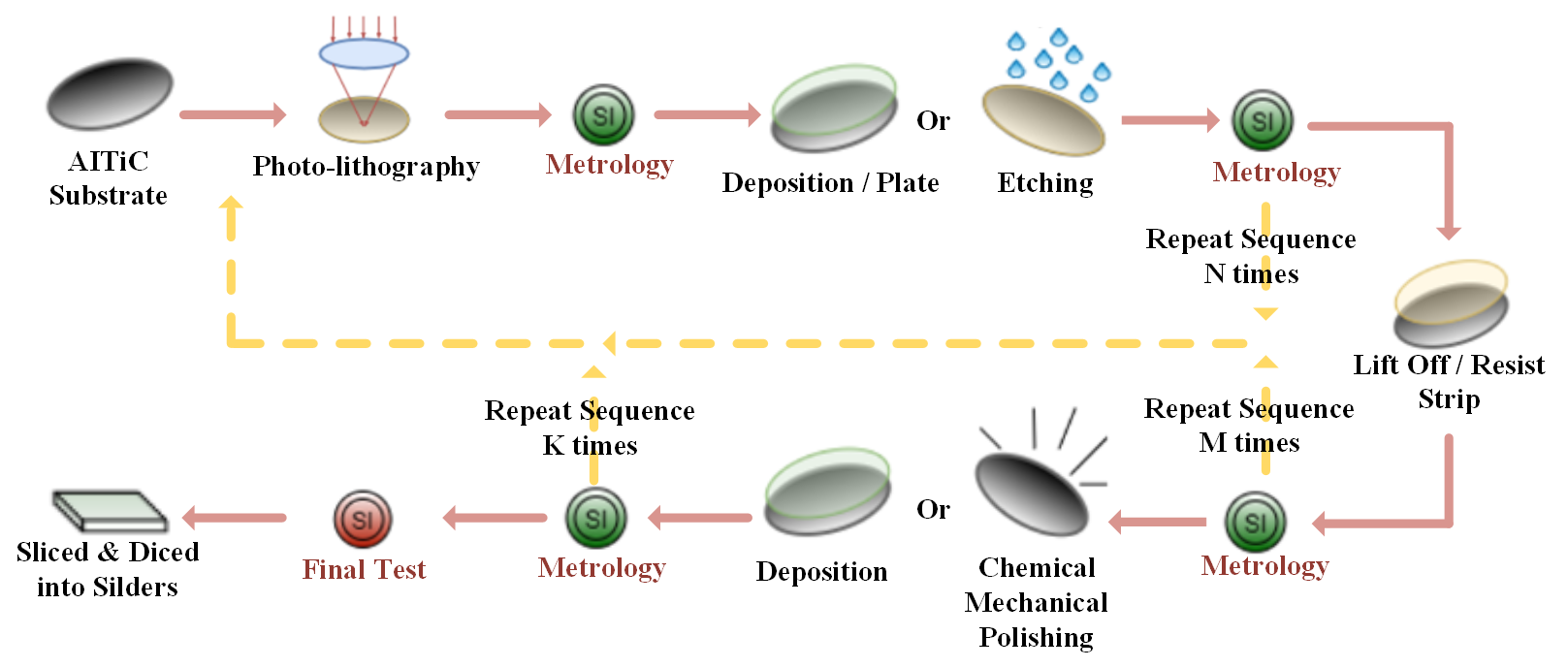}
	\caption{A high-level simplified overview of wafer manufacturing process}
	\label{Fig:wafer}
\end{figure}

In semiconductor manufacturing, it is highly desirable to develop the soft-sensing-based system for defective wafer detection\cite{fan2020defective}. There are two main approaches to establish soft sensing models on defective wafer detection: the model-based and data-driven approaches \cite{sun2021survey}. The model-based approach requires a wealth of experience and knowledge to establish the mathematical models and describe the equipment conditions. Fig. \ref{Fig:wafer} depicts a high level, simplified wafer manufacturing process. The complex structure of today's wafer production equipment and control monitoring systems make it difficult to solve the time-varying unexpected disturbances and uncertainties with model-based approaches. As a result, data-driven modeling has become more prevalent in defective wafer detection \cite{rostami2018automatic}. In the last two decades, deep learning has emerged as a promising soft-sensing approach to process complex data and to learn representations of data with multiple levels of abstract information \cite{yan2020deep}. Soft-sensing technology prevents yield loss and reduce the processing cycle time by effectively monitoring the production quality. With the complex valued neural networks, deep learning model can extract the temporal and nonlinear features and find feature mappings between multiple predicted and observable monitoring variables and wafer quality.

However, despite the success of deep learning models in soft-sensing system, the underlying logic of the deep learning framework is considered as a black-box and hard to understand. Before deploying the deep learning model to monitor the quality of wafer in the industry, it is necessary to understand the model's operation and how input data results in a specific decision. To investigate how the deep learning model works, the deep visualization is proposed \cite{yosinski2015understanding,psuj2018multi}. The deep visualization is employed to explore the knowledge on inner workings of the deep learning models and to understand the focus of intermediate features over given data. The deep visualization provides the interpretation of feature mapping and the perception of how the monitoring sensor data influence a specific model prediction\cite{simonyan2013deep}. The interpretation of deep visualization makes an effective observation of how the deep learning model works, which makes it more intuitive to both manufacturing engineers and researchers. Moreover, the results obtained from deep visualization can be exploited to further optimize the deep learning model. 

In this paper, we propose a soft-sensing system for defective wafer detection based on the deep learning approach and apply the deep visualization for model optimization. We summary the contribution as follows:

\begin{enumerate}
    \item The deep learning model based on dilated convolutional neural network (CNN) is proposed for the defective wafer detection on highly imbalanced monitoring sensor dataset. 
    
    \item The deep visualization is applied to the system for understanding the operating mechanism behind the deep learning model. The contributions of the sensor data are recognized and are further utilized for fine-tuning the model.
    
    \item Experiments are performed based on deep visualization to demonstrate the feasibility and effectiveness of fine-tuning the model guided by the interpretation of deep visualization.
\end{enumerate}

The paper is organized as follows. Section \ref{sec:related_work} reviews the related work. Section \ref{Sec:system} presents the deep learning based soft sensing system and how to fine-tune the model with deep visualization. Section \ref{Sec:exper} illustrates the experiments with analysis. Finally, section \ref{Sec:dis} makes up a discussion and section \ref{Sec:con} concludes the paper. 

\section{Related Work} \label{sec:related_work}
\subsection{Data-Driven Soft Sensors Models}

As smart sensors with sophisticated hardware evolve, they generate a large amount of valuable monitoring variables, which indicate the status of the manufacturing processes and can be applied in data-driven soft-sensing models. The data-driven soft sensor models include partial least squares (PLS) regression \cite{zheng2018semisupervised}, support vector machine (SVM) \cite{yan2004soft}, Random Forest \cite{wang2020random}, k-nearest neighbors(k-NN) \cite{cheng2018monitoring} and deep learning (DL) \cite{yao2017deep}. In recent years, the soft-sensing models based on deep learning approach have been developed and have proven to be effective. Deep learning approaches have developed rapidly and have become prevalent in big complex data processing \cite{jiang2019distributed,huang2019adversarial}. Besides, the emergence of deep learning techniques reduces the need of domain specialists for feature extraction \cite{ahn2020deep,qian2020wearable,huang2020reliable}. Yuan et al. \cite{yuan2020supervised} applied the deep belief network (DBN) to learn the dynamics and various local correlations of different variable combinations for quality prediction. Wang et al. \cite{wang2019dynamic} used two Convolutional neural network (CNN)-based soft sensor models to deal with abundant processing variables. Zhu et al. \cite{zhu2018deep} utilized the CNN-based model to extract the time-independent correlations among the monitoring variables. Wei et al. \cite{wei2015soft} transferred the raw data into the frequency domain and enabled the CNNs to learn high invariance according to signal translation, scaling and distortion. 

\subsection{Deep Visualization}

In recent decades, deep neural network (DNN-based) models have shown unprecedented performance in soft-sensing system. However, the architectural complexity and high nonlinearities of the DNNs model make comprehension difficult. Before deploying a well-trained DNNs model to the industry, it is vital to be able to interpret and explain the mechanisms of the model. Deep visualization is proposed to build the trust in the model by evaluating its output. Deep visualization can be regarded as an efficient way to open and explain the ``black-box'' of DNNs \cite{qian2019smart}. The deep visualization has been successfully applied to computer vision and natural language processing (NLP), but rarely utilized in soft-sensing system. Simonyan et al. \cite{simonyan2013deep} applied saliency maps to address the visualization of image classification models. Yosinski et al. \cite{yosinski2015understanding} proposed an approach to visualize the activation of each layer in the trained models. Selvaraju et al. \cite{selvaraju2017grad} proposed an approach named Grad-CAM based on gradient-weighted class activation mapping for image visualization. To provide better human-interpretive visual explanations, researchers proposed the Grad-CAM++ \cite{chattopadhay2018grad}.

\subsection{Saliency Map-Based Deep Visualization}
To understand the deep learning-based classification model, the saliency map-based visualization approach is introduced. The method translates the information into a comprehensible context. Let $C$ be the number of classes of the classification head. Given an input of sensor features $I$, and the class $c \in C$ to be visualized. To get the rank of the sensor features of $I_0 \in I$ according to the influence on the output of class $t$, the class saliency visualisation based approaches \cite{simonyan2013deep} are applied. Take the linear score model as the example, and the score for the model class $t$ is computed by the first-order Taylor expansion as shown in Equation \ref{score_C}:

\begin{equation}\label{score_C}
    S_c(I) \approx \omega^T I + b
\end{equation}
To detect which sensor features change the least and contribute to the class score the most, the derivatives of $S_t$ with the input sensor features of $I_0 \in I$ are applied as shown in Equation \ref{d_w}.

\begin{equation}\label{d_w}
    \nabla I_0 = \frac{\partial S_c}{\partial I} \bigg|_{I = I_0} 
\end{equation}

Generally, given an input $I$ and a class $c$, the score $S_t$ of the class $c$ can be calculated among the models. And the derivative of $S_t$ represents the features contribution to that specific class $c$.

\section{Methodology} 
\label{Sec:system}
\begin{figure*}[ht]
	\centering
	\includegraphics[width=0.98\textwidth]{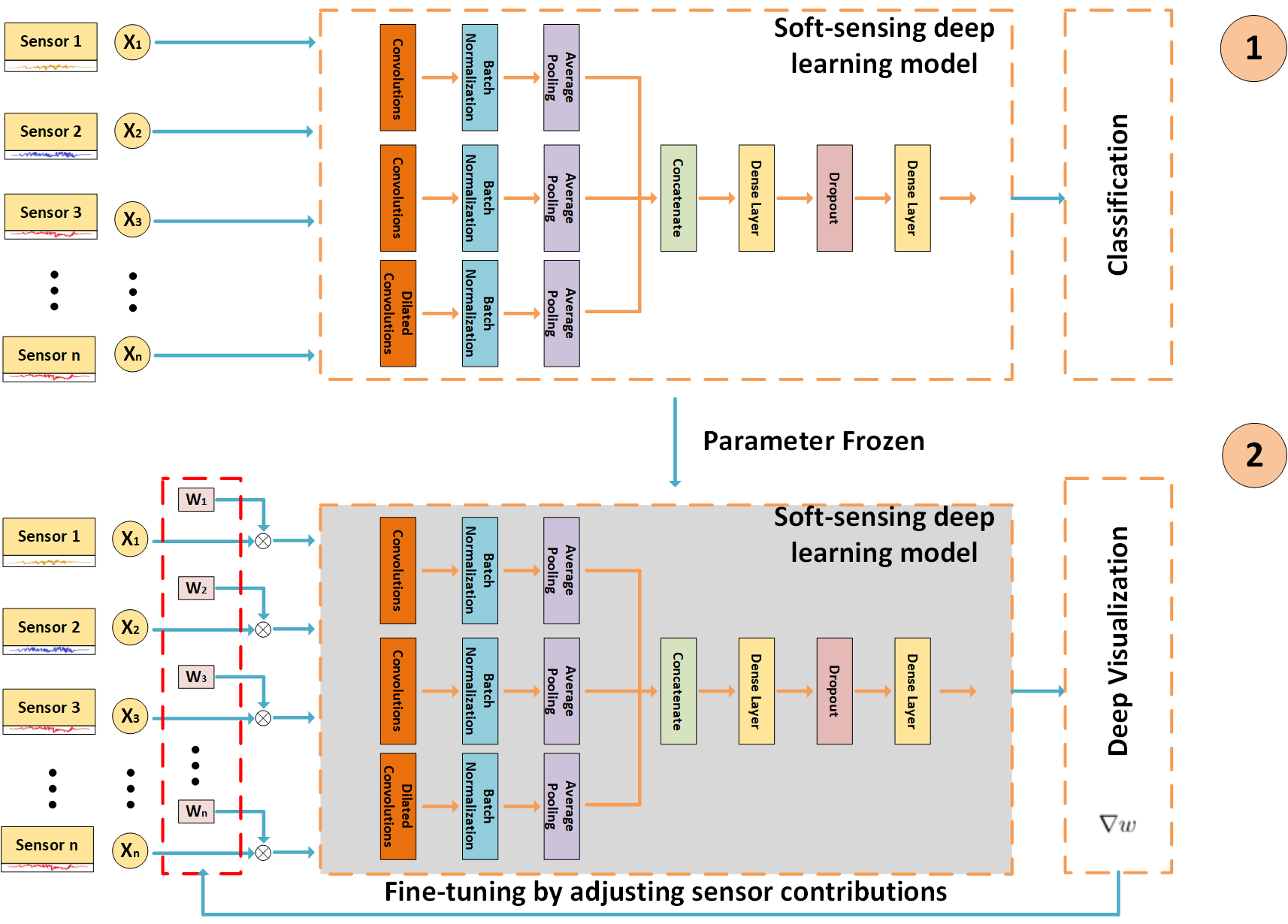}
	\caption{This figure shows the working flowchart of our system. Input of our system is the sensor data and output is the wafer measurement failure. At the beginning, the model initiates the training process from scratch to detect the defective wafer. After well-training the model, the deep visualization is applied to fine-tune the model by decreasing the importance of features which lead the model to misclassify wafer failure.}
	\label{Fig:flowchart}
\end{figure*}

In this section, the CNN-based soft-sensing system for detecting wafer quality failure is introduced. As shown in Fig. \ref{Fig:flowchart}, we demonstrate the work flow of the proposed system. The multivariate time-series sensor data are fed into the soft-sensing model for defective wafer prediction. At the first training stage, the model initiates the training process to classify the defective wafer. At the second training stage, after training the model, the deep visualization approach is utilized to fine-tune the model with those misclassified samples. The assumption behind the proposed approach is that the model can be fine-tuned by decreasing the importance of features which lead the model to misclassify wafer defects. 

The technical details of training the deep learning-based soft-sensing system are proposed in \ref{sec:m-dp_model}. How to fine-tune the model according to the deep visualization is discussed in \ref{sec:m-finetune}.

\subsection{Deep Learning-based Soft-sensing System} \label{sec:m-dp_model}

Deep learning based soft sensing has been applied to soft sensing systems due to its promising performance on the classification task. To process the multi-variate time-series data, the convolutional layers are performed by applying convolutional kernels and sliding them over the time series to extract features through both temporal and spatial dimension. Each convolutional layer contains 32 kernels with kernel size of $(2 \times 1)$, $(2 \times 3)$ and $(2 \times 5)$ respectively. To increase the perceptive field without increasing the model computations, the traditional convolution kernels are replaced by dilated convolution kernels. Specifically, three convolutional layers are applied in parallel to extract the features of the input data, one of the traditional convolutional kernels is replaced by dilated kernel. The output of $i^{th}$ the dilated convolution at layer $l$ can be formulated as in Eq. \ref{dilated_cnn}. $d$ means the dilation rate and $k$ is the convolution kernel size. 

\begin{equation}\label{dilated_cnn}
    Y_{i}^{l} = \sum W_i^l x_{i-k\times d}^{l-1} + b^l
\end{equation}

In the proposed model, the output of the convolutional layer is regularized through applying the batch normalization. Besides, the average pooling layer is applied after the convolution structure to create the down-sampled feature map, which makes the extracted feature maps to be more closely to the classification categories and reduces the tendency of overfitting. The outputs from three parallel convolutional blocks are concatenated. The ensemble approximation of three blocks can overfit the training data, so that the dropout layer with the rate of $0.5$ is applied to the model for reducing the effect of the model learning the statistical noise of the training data. The activation function is further utilized in the proposed model. The design enables the system to make full use of the large amount of data effectively. Coupled with the deep learning architecture, the model is able to learn the nonlinearity of the complex input monitoring sensor data. To address the high imbalance problem in the dataset, the weighted binary cross-entropy for all classes is defined in Equation \ref{weighted_loss}.

\begin{equation}\label{weighted_loss}
    L_i(y_i,\hat{y}_i) = -(\beta y_i log(\hat{y}_i) + (1-y_i) log(1-\hat{y}_i))
\end{equation}

Where $\beta$ is the parameter for weighting the classes to deal with the imbalance problem, $y_i$ and $\hat{y_i}$ are the ground truth and model predictions, respectively.

\subsection{Fine-tuning Algorithm with Deep Visualization} \label{sec:m-finetune}

To improve the performance of the model for wafer defects, we propose an algorithm to realize the closed-loop model fine-tuned by the deep visualization. The parameters in the well-trained deep learning model are frozen and the deep visualization is applied to the model. Deep visualization provides the explanation about feature mapping and the perception of how the input data influence the model predictions. We incorporate the insights gathered from deep visualization to the model and allow the learned information to guide the fine-tuning processing. As shown in Fig. \ref{Fig:flowchart}, a weights layer is designed to adjust the importance for each input sensor data before feeding them into the model. The deep visualization based on the saliency map indicates the high correlation of sensor features with model prediction so that we can take advantage of such correlation to fine-tune the model by adjusting the weights for each input data. We denote $I^{(n)}$ as the input of layer $L^{(n)}$, where $n \in [1,...,N]$ represents the layer index in a N layer neural network model. To fine-tune the model, the sensor weight $\omega_1^1, ... \omega_i^1$ are designed to each data, where $i \in I_0$ is the index of data and we add this weights to the very first layer of the model (superscript is 1). According to the definition of Equation \ref{score_C}, the class score $S_t(I)$ can be redefined as shown in the Equation \ref{score_C2}, where the $w^1$ is the weights to adjust the sensor contributions, $I$ is the input of the sensor data. 

\begin{equation}\label{score_C2}
    S_t(I) \approx \omega^T (\omega^1 \cdot I) + b
\end{equation}

Recalling the chain rule, the gradient of model output of a with the input $I_0$ is calculated by the Equation \ref{d_w2}.

\begin{equation}\label{d_w2}
    \nabla I_0 = \frac{\partial I^1}{\partial I} \frac{\partial S_t}{\partial I^1} = w^1 \cdot \frac{\partial S_t}{\partial I^1}\bigg|_{I = I_0}
\end{equation}

The equation \ref{w_update} is applied to update the sensor weights, where $\alpha$ is learning rate.

\begin{equation}\label{w_update}
    w^1_{updated} = w^1 - sign(\alpha \times \nabla w^1)
\end{equation}

Where $\nabla w^1$ is defined in Equation \ref{w_update1}.

\begin{equation}\label{w_update1}
    \nabla w^1 =\frac{\partial I^1}{\partial w^1} \frac{\partial S_t}{\partial I^1} = I \cdot \frac{\partial S_t}{\partial I^1} \bigg|_{I = I_0}
\end{equation}

According to the Equation \ref{w_update1}, the weights of each sensor are adjusted according to the deep visualization result. It is noted that $ \frac{\partial S_t}{\partial I^1}$ is the saliency map derived by visualizing the trained model. 

To deal with the high-imbalance data problem, the standardization is applied to the saliency feature map for each class, as shown in Equation \ref{w_norm}.
\begin{equation}\label{w_norm}
    \hat{X_i} = 
    \begin{cases}
        \frac{X_i-\mu}{\sigma} & X_i\neq0\\
        0 & X_i=0
    \end{cases}
\end{equation}

Where $\mu$ is the mean of the variables in featuring mapping and the $\sigma$ is the unit standard deviation. The derivatives of $S_t$ (saliency map) of samples which are predicted as class $c$ by the model are averaged to update the the sensor weights at the first layer of the model $\omega_1^1 ... \omega_i^1$. Besides, we clip each variable in the saliency map with small contributions to the weight update to get a better convergence. We take the absolute value by computing $|W^1-\nabla w^1| \leq \theta$ in either direction, positive and negative. We prefer to clip the features so that the updates do not take large steps outside the region where the model approximation is most valid. 

The key to improve the model performance is to derive insights from the deep visualization about why the model ends up misclassifying. It has the potential opportunity for improving model performance by concentrating more on those misclassified samples rather than those well-classified samples. In this section, we describe how to fine-tuning deep learning model by adjusting the importance of sensors before feeding into the model based on the misclassified samples including False Positive (FP) and False Negative (FN) samples. Specifically, the FP means the output where the model predicts the positive class by mistake, and FN represents the output where the model predicts the negative class to be the positive. For each misclassified samples, the saliency map indicates which sensor data contribute more to the wrong class. To fine-tune the model, we minimize the effects of those sensor data. The weights of the sensor data are decreased according to equation \ref{w_update} and the sign is positive. Similarly, the feature contributions to the correct class are calculated and are expected to increase, so that the weights are updated according to equation \ref{w_update} and the sign is assigned to be negative. Note that, the misclassified samples have scores for both the correct class and wrong classes. According to the algorithm, the effect of data that leads the model to detect correctly increases and vice versa, which can assist the model to eliminate the error of prediction.

\section{Experiments and Results} \label{Sec:exper}
\subsection{Seagate Soft-sensing Data}

\begin{table}[ht]
\renewcommand{\arraystretch}{1.3}
\caption{Summary of the datasets with number of positive and negative samples for each Key Quality Indicator (KQI) task.}
\label{tbl:data}
\centering
\setlength\tabcolsep{7.5pt}
\begin{tabular}{|c|c|c|c|c|c|c|}
\hline
 & \multicolumn{2}{|c|}{Train} & \multicolumn{2}{|c|}{Valid} & \multicolumn{2}{|c|}{Test}\\
\hline
Task & Neg & Pos & Neg & Pos & Neg & Pos\\
\hline
KQI-1 & 6020 & 272 & 1417 & 13 & 878 & 10\\
KQI-2 & 10288 & 33 & 1509 & 5 & 950 & 2\\
KQI-3 & 42989 & 200 & 7795 & 43 & 5414 & 48\\
KQI-4 & 11114 & 132 & 1594 & 23 & 1989 & 33\\
KQI-5 & 32794 & 428 & 4283 & 91 & 3567 & 49\\
KQI-6 & 64007 & 709 & 11833 & 68 &  9123 & 86\\
KQI-7 & 117332 & 1702 & 19663 & 482 & 16975 & 371\\
KQI-8 & 1748 & 443 & 196 & 39 & 975 & 8\\
KQI-9 & 22420 & 86 & 4225 & 6 & 2906 & 12\\
KQI-10 & 7874 & 48 & 1788 & 4 & 1151 & 5\\
KQI-11 & 35874 & 227 & 6231 & 36 & 5114 & 43\\
\hline
\multicolumn{7}{c}{*Neg/Pos: the passed/failed samples of corresponding key indicator.}
\end{tabular}
\end{table}

The datasets are collected from Seagate wafer manufacturing factories in both the US and Ireland. The high dimensional time-series sensor data are generated from different manufacturing machines. As shown in Fig. \ref{Fig:wafer}, a high-level wafer processing flow is illustrated. In the wafer manufacturing process, an AlTIC wafer is processed through several stages including deposition, coating, lithography, etching and polishing. To monitor and control the quality of the wafer to meet the desired yield, these processes are measured by several measurement tools. These processes are highly complex and sensitive to both incoming and point of process effects. The measurement function plays a significant role in the quality control. For each measurement, there is a pass or fail decision based on the the numerical measurement values. Each time-series multivariate sample is mapped to 11 measurement tasks, which are summarized in Table \ref{tbl:data}. The sensor data are slided by two time steps so that there are two time-step data in each sample.

\subsection{Evaluation Metric}
To evaluate the effectiveness of the proposed system, the area under the receiver operating characteristic (AUROC) is reported. Besides, TPR (recall) is applied to validate the model performance. In the manufacturing industry soft-sensing system recall is an essential indicator to prevent passing bad quality wafers to the next processing step.

\subsection{Results}
\subsubsection{Deep Visualization}

\begin{figure*}[ht]
	\centering
	\includegraphics[width=0.98\textwidth]{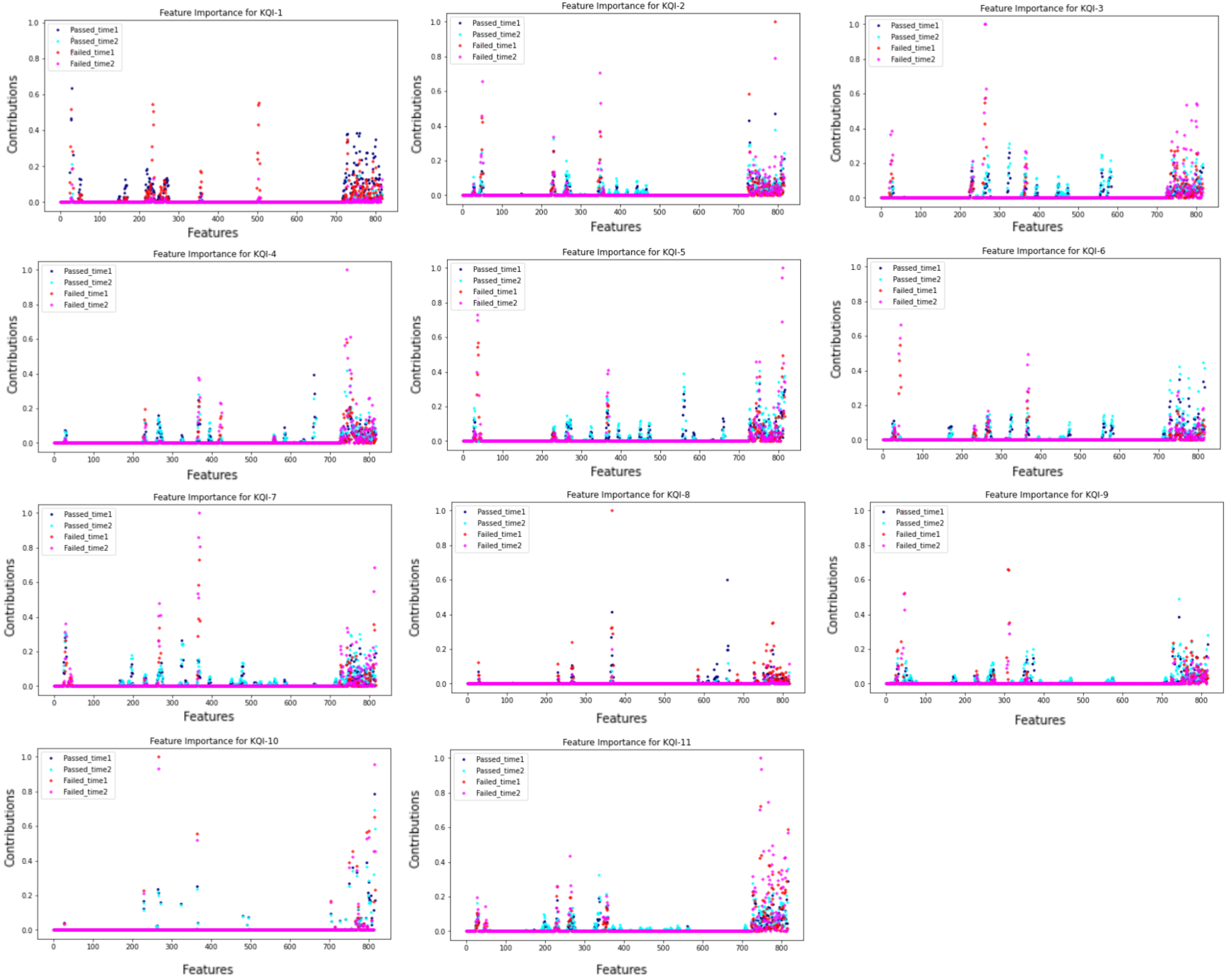}
	\caption{The feature contributions through the model among two time steps with two classes (passed or failed). Y-axis is the feature importance and X-axis is the $i^{th}$ feature. The passed\_time1 and passed\_time2 represents the feature contributions got from deep visualization to the model class ``passed'' with two time steps separately. The Failed\_time1 and Failed\_time2 are the results got from model class ``failed'' with two time steps.}
	\label{Fig:all_tps}
\end{figure*}

As shown in Fig. \ref{Fig:all_tps}, the contribution of sensor data through the CNN model to failed and passed classes is illustrated. The saliency map are derived from the deep visualization algorithm, the average results through all samples for each measurement task are shown. Among 11 different measurement tasks, it is worth to note that the feature contribution of the CNN model for predicting the samples as pass and failure is dissimilar. The results explain how the CNN model works and show the distribution of sensor contributions. With one sample, the feature contribution to the model can be slightly different through time dimension according to the different extracted temporal information. In general, an important sensor can contribute more to the model, but the contributions are inconsistent among the classes. The deep visualization not only provides the explanation of CNN model, but also has the potential to limit the number of active sensors needing to be monitored and controlled.

\begin{figure*}[ht]
	\centering
	\includegraphics[width=0.99\textwidth]{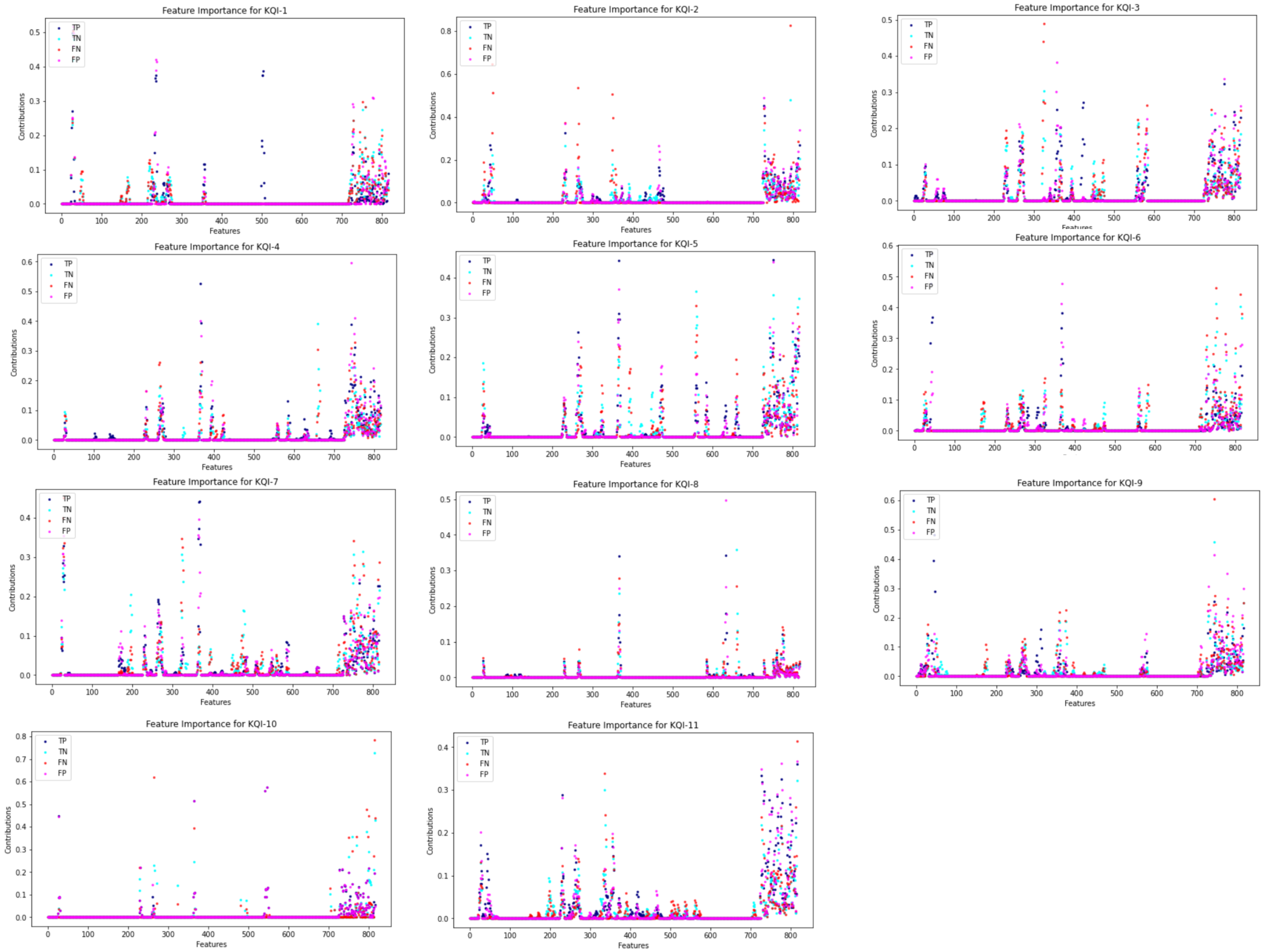}
	\caption{The feature contributions through the model among TP, FP, TN, and FN samples.}
	\label{Fig:tps}
\end{figure*}

In order to further understand the model, the sensor contributions of true positive (TP), false positive (FP), true negative (TN), and false negative (FN) samples are demonstrated, as shown in Fig. \ref{Fig:tps}. In this paper, TP is the output where the model recognizes the defective wafer as ``failed''. FP means the output where the model predicts the ``good'' wafer to be defected. TN is the number of samples the model successfully detected as ``good''. FN represents the output where the model predicts the ``failed'' wafer to be passed. The sensor contributions are dissimilar between TP and FN, as well as the FP and TN. We calculate the averaged sensor contributions for each sample. Specifically, the sensor contributions of one sample are averaged through the time dimension and through all classes. It is worth to note that the data importance of TP, TN, FP, and FN samples are different. Combined the Fig. \ref{Fig:all_tps} with Fig. \ref{Fig:tps}, we could find that it has correlation between the feature contribution to Failed class and the TP feature contribution, and the correlation between passed class and the TN. Motivated by the deep visualization results, we apply the weight layer to decrease the sensor contributions of the misclassified samples, which can be further applied for improving the system performance.

\subsubsection{Model Finetuned with Deep Visualization}
In this experiment, we validate the performance of the proposed CNN-model and the fine-tuned CNN-model on deep visualization based on the Seagate dataset. The quantitative results are shown in Table \ref{tbl:auc}. The AUROC is calculated. The performance of all 11 tasks from fine-tuned model outperforms the CNN-model on AUROC. The results have shown that fine-tuning the model with misclassified samples (FP and FN) increases the model performance. We decrease the weights of those features of misclassified samples that lead the model to the wrong class and increasing the weights of the features that contribute more to the correct class.

\begin{table}[ht]
\renewcommand{\arraystretch}{1.3}
\caption{AUROC score of finetuned CNN and Baselines on test dataset. CNN is the baseline model and $CNN_{dv}$ is the CNN model fine-tuned by decreasing the feature contributions of misclassified samples. }
\label{tbl:auc}
\centering
\setlength\tabcolsep{8pt}
\begin{tabular}{|c|c|c|}
\hline
Model & $CNN$ & $CNN_{dv}$\\
\hline
KQI-1 & 0.8441 & \textbf{0.8446}\\
KQI-2 & 0.7753 & \textbf{0.7805}\\
KQI-3 & 0.7885 & \textbf{0.7943}\\
KQI-4 & 0.8545 & \textbf{0.8567}\\
KQI-5 & 0.6145 & \textbf{0.6228}\\
KQI-6 & 0.5704 & \textbf{0.585}\\
KQI-7 & 0.6679 & \textbf{0.6689}\\
KQI-8 & 0.6738 & \textbf{0.6856}\\
KQI-9 & 0.833 & \textbf{0.8331}\\
KQI-10 & 0.7976 & \textbf{0.799}\\
KQI-11 & 0.783 & \textbf{0.7863}\\
\hline
\end{tabular}
\end{table}

\section{Discussion} \label{Sec:dis}

\subsection{Can we Improve Recall by Deep Visualization? }

Recall is a crucial statistic indicator to assess how many of the true positives were recalled or found. As mentioned above, recall can be very important in wafer manufacturing industries to ensure no defective wafer moves to the next stage in the build. In the scenario when the majority samples in the dataset are negative, recall is an important indicator to validate the effectiveness of the model. In this part, we fine-tune the model by assigning more weights to FN samples than FP samples to improve the recall. We remove the measurements if the whole test dataset contains less than 10 positive samples (wafer is regarded as failed). As shown in Table. \ref{tbl:data_tp}, the recall increases significantly with similar model performance (AUROC). The results have shown that deep visualization provides a promising way to improve the recall of the system. To adjust the sensor contributions for those weighted misclassfied samples, the performance of the system can be improved.

\begin{table}[ht]
\renewcommand{\arraystretch}{1.3}
\caption{AUROC score and recall of finetuned CNN and Baselines on test dataset. CNN is the baseline model and $CNN_{dv}$ is the CNN model fine-tuned by decreasing the feature contributions of FN samples.}
\label{tbl:data_tp}
\centering
\setlength\tabcolsep{7.5pt}
\begin{tabular}{|c|c|c|c|c|c|c|}
\hline
 & \multicolumn{2}{|c|}{$CNN$} & \multicolumn{2}{|c|}{$CNN_{dv}$}\\
\hline
Task & TPR & AUROC & TPR & AUROC\\
\hline
KQI-3 & 0.7917 & \textbf{0.7885} & \textbf{0.8125}&0.7776\\
KQI-4 & 0.3333 & \textbf{0.8545} & \textbf{0.9091} & 0.8246\\
KQI-5 & 0.2245 & 0.6145 & \textbf{0.5714} & \textbf{0.6166}\\
KQI-6 & 0.3372 & 0.5704 & \textbf{0.5116} & \textbf{0.5955} \\
KQI-7 & 0.5174 & \textbf{0.6679} & \textbf{0.6404} &0.6668\\
KQI-9 & 0.75 & \textbf{0.833} & \textbf{0.8333} & 0.8275\\
KQI-11 & 0.3953 & \textbf{0.783} & \textbf{0.6744} & 0.7606\\
\hline
\end{tabular}
\end{table}

\subsection{What can we do with deep visualization?}
In our proposed system, the deep visualization is applied and utilized for optimizing the system. Traditional deep visualization method provides the interpretation for the deep learning model. Rather than simply interpreting the model, it is more meaningful to exploit the knowledge learned from deep visualization to improve the model performance. In this study, we are not merely finding the way to open the deep learning 'black box', we are further utilizing the deep visualization to fine-tune the deep learning model. This domain should be a more valuable issue to be discussed and be concerned about. 

\section{Conclusion} \label{Sec:con}
In this paper, we build a deep learning model for defective wafer detection and apply deep visualization to understand the underlying logic about how the time-series soft-sensing data affect the model prediction. The result of deep visualization is further utilized to fine-tune the model based on the weight layer to adjust the sensor contributions of the input data. The proposed fine-tuning method is to minimize the effects of the input sensors that lead the model to misclassify. The experiments based on Seagate wafer dataset are performed to further validate the effectiveness of the entire system. Both quantitative and qualitative results demonstrate that the proposed system outperforms the state of art soft sensing model.

\section*{Acknowledgments}

The authors would like to thank Seagate Technology for the support of this study, the Seagate Lyve Cloud team for providing the data infrastructure, and the Seagate Open Source Program Office for open sourcing data sets.  Special thanks to the Seagate Data Analytics and Reporting Systems team for inspiring the discussions.

\bibliographystyle{IEEEtran}
\bibliography{./References/reference}

\end{document}